\documentclass[runningheads,11pt]{llncs}

\usepackage{amsmath,amssymb,amsfonts}
\usepackage{graphicx,xcolor}
\usepackage{booktabs}
\usepackage{algorithm}
\usepackage{algpseudocode}
\usepackage{microtype}
\usepackage[hidelinks]{hyperref}
\usepackage{enumerate}

\usepackage{geometry}
\begin{document}

\title{Modular Jets for Supervised Pipelines: \\ Diagnosing Mirage vs Identifiability}

\titlerunning{Modular Jets for Supervised Pipelines}

\author{Suman Sanyal}
\authorrunning{S. Sanyal}
\institute{Goa Institute of Management, Goa, India \\ 
\email{sanyal@gim.ac.in}}

\maketitle

\begin{abstract}
Classical supervised learning evaluates models primarily via predictive risk on hold-out data. Such evaluations quantify how well a function behaves on a distribution, but they do not address whether the internal decomposition of a model is uniquely determined by the data and evaluation design. In this paper, we introduce \emph{Modular Jets} for regression and classification pipelines. Given a task manifold (input space), a modular decomposition, and access to module-level representations, we estimate empirical jets, which are local linear response maps that describe how each module reacts to small structured perturbations of the input. We propose an empirical notion of \emph{mirage} regimes, where multiple distinct modular decompositions induce indistinguishable jets and thus remain observationally equivalent, and contrast this with an \emph{identifiable} regime, where the observed jets single out a decomposition up to natural symmetries. In the setting of two-module linear regression pipelines we prove a jet-identifiability theorem. Under mild rank assumptions and access to module-level jets, the internal factorisation is uniquely determined, whereas risk-only evaluation admits a large family of mirage decompositions that implement the same input-to-output map. We then present an algorithm (MoJet) for empirical jet estimation and mirage diagnostics, and illustrate the framework using linear and deep regression as well as pipeline classification.
\keywords{Identifiability \and Mirage \and Modular Jets \and Interpretability \and Model evaluation}
\end{abstract}

\section{Introduction}
\label{sec:intro}

Supervised learning systems are almost universally evaluated by their predictive performance on held-out data, like mean-squared error for regression, accuracy or AUC for classification, and so on. These metrics estimate the risk of a learned function with respect to an underlying data distribution, and they are indispensable for model selection and comparison~\cite{vapnik1998statistical,hastie2009elements}. At the same time, recent work on causal and disentangled representation learning emphasises that low risk by itself need not pin down a unique internal decomposition or causal structure of a model~\cite{scholkopf2021toward}. Moreover, classical risk-based evaluation leaves an important question unaddressed. \emph{What, if anything, do risk-based evaluations tell us about the internal structure of a model?} In modern practice, even simple regression or classification pipelines are composed of multiple modules like preprocessing, feature extraction, dimensionality reduction, representation learning, prediction heads, and calibration layers. Practitioners often tell a semantic story about such pipelines (for example, ``this bottleneck encodes factors A and B, while the head combines them into the label''), but standard evaluation does not test whether these stories are uniquely enforced by the data, or whether many different internal mechanisms would behave identically on the observed distribution.

In this paper, we propose \emph{Modular Jets} as a simple tool to bridge this gap in the classical supervised setting. The idea is to treat a learned system as a composition of modules and to study its behaviour on a \emph{task manifold}, which in the present setting we take to be a subset of the input space. Around each input $x_0$, we probe the model with a small neighbourhood of structured perturbations $x_0 + \delta$, and we record the responses not only at the output but also at taps into intermediate modules. For each module, we fit a local linear model that maps input perturbations to changes in its tapped representation. This local pair (representation at $x_0$, empirical Jacobian) is the module's \emph{jet} at $x_0$.

We then ask an identifiability-style question. \emph{Given a class of modular decompositions and a perturbation design, are the observed jets sufficient to single out a decomposition (up to natural equivalences), or can multiple decompositions induce virtually indistinguishable jets on all probed inputs?} We refer to the latter situation as a \emph{mirage regime} where risk-based evaluations and local response patterns under the chosen perturbations cannot distinguish between different internal stories. Conversely, when different decompositions are empirically separated by their jets, we say that the evaluation design provides evidence for decomposition identifiability. Our contributions are both conceptual and methodological.
\begin{enumerate}
    \item We introduce Modular Jets in the simplest possible setting of classical regression and classification pipelines, with the input space as the task manifold.
    \item We introduce an empirical notion of mirage versus identifiability for modular decompositions, distinct from but complementary to standard test risk.
    \item We prove a jet-identifiability theorem for linear two-module regression pipelines, together with an explicit $r^2$-parameter family of mirage factorizations that are invisible to risk-only evaluation.
    \item We provide the Modular Jet Diagnostics (MoJet) algorithm for empirical jet estimation and mirage diagnostics, based on local perturbations and ridge-regularised linear fits.
    \item We present experimental setups in linear and deep regression, as well as in pipeline classification, where models with identical risk exhibit different behavior under jet-based analysis.
\end{enumerate}
Our aim is to make the identifiability question, often discussed in the context of latent variable models and deep representation learning, accessible in a familiar supervised-learning setting, and to invite further work on principled evaluation designs that constrain not only performance but also internal structure. The rest of the paper is organised as follows. Section~\ref{sec:related-work} reviews connections to interpretability, causal representation learning, and mechanistic analysis of large models. Section~\ref{sec:setup} sets up the supervised-learning framework and the notion of modular pipelines that we study. Section~\ref{sec:jets} introduces empirical modular jets, explains how local linear response maps are estimated from perturbations, and, in subsection~\ref{subsec:jet-identifiability-linear}, we present a jet-identifiability theorem for linear two-module pipelines together with a mirage family that is invisible to risk-only evaluation. Section~\ref{sec:algorithm} then presents MoJet, a model-agnostic algorithm for estimating jets and computing rank- and similarity-based diagnostics. Section~\ref{sec:experiments} illustrates the approach on a series of synthetic and real-data examples, including linear regression, a two-module deep regressor, and a digits classification task comparing a PCA and logistic pipeline to an MLP, and reports small robustness and compute-cost ablations for the digits setting. Finally, in Section~\ref{sec:conclusion}, we discuss limitations, connections to existing work on interpretability and invariance, and directions for extending Modular Jets to richer model classes and evaluation designs.

\section{Related work}
\label{sec:related-work}

Our work sits at the intersection of four strands. (i) Post-hoc interpretability and model auditing, (ii) causal and disentangled representation learning with identifiability guarantees, (iii) self-supervised and contrastive learning viewed through identifiability, and (iv) emerging mechanistic analyses and evaluation protocols for large models.

\paragraph{Interpretability and model auditing.}
There is a substantial literature on interpretability in classical and deep learning, ranging from post-hoc explanations (feature importance, saliency maps, example-based explanations) to inherently interpretable models~\cite{goodfellow2016deep,doshi2017towards,rudin2019stop}. Most of this work focuses on making the behaviour of a \emph{given} trained model more understandable to humans, often without asking whether the model’s internal decomposition is uniquely determined by the data and evaluation design~\cite{holzinger2020explainable,gilpin2018explaining,marcinkevivcs2023interpretable,li2022interpretable,csahin2025unlocking}. Modular Jets are complementary. They do not try to enforce semantic factors by construction, but instead ask which aspects of a chosen modular decomposition are in principle constrained by local response behaviour under perturbations.

\paragraph{Identifiability in causal and disentangled representation learning.}
A large body of work studies when latent factors and their causal structure can be recovered (up to symmetries) from observational and interventional data~\cite{scholkopf2021toward,vonkugelgen2021nonparametric}. Negative results show that, without inductive biases or supervision, disentangled factors are generally \emph{not} identifiable from i.i.d.\ observations alone~\cite{locatello2019challenging}. Positive results introduce temporal structure, auxiliary variables, or interventions, for example in nonlinear ICA via time-contrastive learning~\cite{hyvarinen2016timecontrastive}, or in multi-view causal representation learning with interventions and content blocks~\cite{yao2024multiview}. These works address \emph{generative} identifiability: given a structural data-generating model, when are its latent variables recoverable? In contrast, we work in a supervised setting and treat the learned predictor as given. Rather than recovering ground-truth generative factors, Modular Jets ask: for a fixed input distribution and perturbation design on the task manifold, which modular decompositions of a predictor are observationally indistinguishable (mirages), and when does module-level jet information collapse this ambiguity?

\paragraph{Self-supervised and contrastive learning as identifiability mechanisms.}
Recent results in self-supervised learning also take an explicitly identifiability-oriented view. For contrastive objectives of InfoNCE type, under suitable assumptions on augmentations and negatives, the learned representation effectively inverts the underlying generative process~\cite{zimmermann2021contrastive}, connecting contrastive learning to nonlinear ICA. Conceptually, these results obtain identifiability by baking structure into the training objective and data pipeline. Our work is orthogonal. We do not assume a particular SSL objective or augmentation scheme, but instead introduce a post-hoc evaluation protocol (jets under perturbations) that can be applied to \emph{any} trained supervised pipeline.

\paragraph{Mechanistic analysis and internal probes.}
There is growing interest in probing and mechanistic analysis of large neural systems, especially language models. Linear and nonlinear probes study how information about labels or latent variables is decodable from intermediate representations, and recent mechanistic-interpretability work examines circuits, features, and invariants inside large models for safety and reliability~\cite{semanticlens2025}. Modular Jets are structurally related in that they instrument intermediate modules and treat the trained system as a black box with taps. However, the object of interest is different. We use jets as local linear response objects to support identifiability-style statements about decompositions, rather than to isolate specific neurons or circuits.

\paragraph{Risk-centric evaluation.}
Standard evaluation practice in supervised learning remains overwhelmingly risk-centric~\cite{vapnik1998statistical,hastie2009elements}. Test risk, calibration metrics, and robustness scores quantify functional performance but say little about whether the internal structure of a pipeline is uniquely pinned down. Our contribution can be viewed as adding a lightweight, model-agnostic layer on top of existing evaluations. By pairing risk with module-level jets under structured perturbations, we obtain diagnostics that distinguish between genuinely different internal geometries and mirage regimes where many pipelines look identical under the chosen evaluation design.

\section{Modular supervised pipelines}
\label{sec:setup}

We work in a standard supervised-learning setting. Inputs $x \in \mathcal{X} \subseteq \mathbb{R}^d$ and labels $y \in \mathcal{Y}$ are drawn from an unknown distribution $P_{X,Y}$. A learner selects a function $f: \mathcal{X} \to \widehat{\mathcal{Y}}$ from a hypothesis class $\mathcal{F}$, with the aim of minimising a suitable risk functional
\[
R(f) := \mathbb{E}\bigl[\ell(f(X),Y)\bigr],
\]
for a loss $\ell$ and under $P_{X,Y}$. Empirically, we approximate $R(f)$ by its estimate on a held-out test set. Although this paper focuses on regression (real-valued $Y$ with squared error) and classification (categorical $Y$ with standard classification losses), the constructions apply more broadly~\cite{vapnik1998statistical,hastie2009elements}. We assume that the learned $f$ is realised as a \emph{modular pipeline}
\begin{equation}
\label{eq:pipeline}
    x \xrightarrow{M_1} z_1 \xrightarrow{M_2} z_2 \xrightarrow{} \cdots \xrightarrow{M_K} z_K = \hat{y},
\end{equation}
where each $M_m$ is a module (like preprocessing, feature map, hidden network block, or prediction head), and $z_m$ is its output or internal representation. The exact granularity of the decomposition can vary. In a linear regression with handcrafted features, $M_1$ may encode feature construction and $M_2$ the linear head; in a deep regressor, $M_1$ might be a feature extractor $h$ and $M_2$ a prediction head $g$; in a pipeline classifier, $M_1$ may be PCA or normalisation and $M_2$ a logistic regression classifier. We treat the input space $\mathcal{X}$ as a \emph{task manifold}. Each input $x$ defines a supervised prediction task ``predict $y$ at $x$", and local perturbations around a base input $x_0$ explore nearby tasks. Our questions are formulated relative to (i) the chosen modular decomposition~\eqref{eq:pipeline} and (ii) a given family of input perturbations.

\section{Empirical modular jets}
\label{sec:jets}

\subsection{Definition and estimation}

Fix a modular pipeline~\eqref{eq:pipeline}, a base input $x_0 \in \mathcal{X}$, and a module index $m$. We assume access to a \emph{tap} into module $M_m$
\[
\mathrm{tap}_m: \mathcal{X} \to \mathbb{R}^{d_m}, \qquad
x \mapsto z_m(x),
\]
which returns a vector representation of the module's state when the pipeline is run on input $x$. In the simplest case, $\mathrm{tap}_m$ outputs the internal vector $z_m$, and more generally, it may output a projection or summary. To study the local behavior of $M_m$ around $x_0$, we consider a collection of small input perturbations $\{\delta_j\}_{j=1}^J \subset \mathbb{R}^d$ and define
\[
x^{(0)} := x_0, \qquad x^{(j)} := x_0 + \delta_j, \quad j=1,\dots,J,
\]
subject to $x^{(j)} \in \mathcal{X}$. Running the full pipeline on each $x^{(j)}$ yields tapped representations $z_m^{(j)} := \mathrm{tap}_m(x^{(j)})$. We then approximate the local dependence of $z_m$ on $x$ via a first-order model
\begin{equation}
\label{eq:local-linear}
    z_m^{(j)} \approx z_m^{(0)} + A_m(x_0)\,\delta_j,\qquad j=1,\dots,J,
\end{equation}
where $A_m(x_0) \in \mathbb{R}^{d_m \times d}$ is an empirical Jacobian of the tap with respect to the input at $x_0$. To fit $A_m(x_0)$, we build a matrix of perturbations
\[
\Delta \in \mathbb{R}^{J \times d},\quad
\Delta_{j,\cdot} := \delta_j^\top,
\]
and a matrix of response differences
\[
Y_m \in \mathbb{R}^{J \times d_m},\quad
Y_{m,j,\cdot} := \bigl(z_m^{(j)} - z_m^{(0)}\bigr)^\top.
\]
We then solve a ridge-regularised least-squares problem
\begin{equation}
\label{eq:ridge}
    A_m(x_0) := \arg\min_{A \in \mathbb{R}^{d_m \times d}} 
         \bigl\|Y_m - \Delta A^\top\bigr\|_F^2 + \lambda \|A\|_F^2,
\end{equation}
with regularisation parameter $\lambda \ge 0$. This admits the closed form
\begin{equation}
\label{eq:jet-closed-form}
    A_m(x_0)^\top = (\Delta^\top \Delta + \lambda I_d)^{-1} \Delta^\top Y_m.
\end{equation}
We call the pair
\begin{equation}\label{eq:mojet}
J_m(x_0) := \bigl(z_m^{(0)}, A_m(x_0)\bigr)
\end{equation}
the \emph{empirical modular jet} of $M_m$ at $x_0$. Repeating this procedure for several base inputs $x^{(s)}$ yields a collection of jets $\{J_m(x^{(s)})\}_{s=1}^S$ describing how the module behaves locally across the input space.

\subsection{Mirage regimes and identifiability}

Given a modular pipeline~\eqref{eq:pipeline}, a family of perturbations, and estimated jets $\{J_m(x^{(s)})\}$, we are interested in whether the observed local behaviour is compatible with multiple distinct modular decompositions. We say that the pipeline is in a \emph{mirage regime} (relative to a model class and perturbation design) if there exist two decompositions
\[
x \xrightarrow{\ M_1,\dots,M_K\ } \hat{y},
\qquad
x \xrightarrow{\ \tilde{M}_1,\dots,\tilde{M}_K\ } \tilde{y},
\]
with identical outputs on all probed inputs ($\hat{y}(x) \approx \tilde{y}(x)$ under the perturbation distribution), such that their module-level jets are also indistinguishable in the sense that
\[
J_m(x^{(s)}) \approx \tilde{J}_m(x^{(s)}),\quad \text{for all } m,s,
\]
up to trivial symmetries (for example, linear reparameterisation of a shared bottleneck). In this case, behaviour on the evaluated distribution, together with local response information, does not distinguish between the two internal stories.

Conversely, we say that the decomposition is empirically \emph{identifiable} (in the Modular Jet sense) if, within the considered model class, any alternative decomposition that matches the observed jets is equivalent to the original one under a clearly specified symmetry group. In this paper, we work with simple diagnostics rather than full identifiability theorems, but the distinction is conceptually useful. Standard risk evaluation tests only whether some function $f$ works well on the distribution, whereas jet-based analysis asks how much of the internal structure is in principle constrained by the data and perturbations.

We deliberately treat these as empirical, evaluation-relative notions. The symbols ``$\approx$" above denote closeness under a chosen metric and tolerance on the finite set of probed inputs and perturbations, rather than exact equality in distribution. In Section~\ref{subsec:jet-identifiability-linear} we show how, in an idealised linear setting with exact jets, this empirical perspective connects to a formal identifiability result. This perspective is closely related to recent work on causal representation learning and identifiability, which asks when latent decompositions are uniquely determined from observational and interventional data~\cite{scholkopf2021toward,vonkugelgen2021nonparametric}.

\subsection{Jet-identifiability (linear two-module pipelines)}
\label{subsec:jet-identifiability-linear}

To make the distinction between mirage and identifiability more concrete, we now consider the simplest setting where jets can, in principle, pin down a decomposition. We restrict attention to linear two-module regression pipelines without bias terms and show that
\begin{enumerate}[(a)]
    \item risk-only evaluation admits a large family of mirage decompositions related by invertible bottleneck reparameterisations, and
    \item access to module-level jets at the bottleneck collapses this family to a single decomposition (up to the trivial identity symmetry), under mild rank assumptions.
\end{enumerate}

We assume a random input $X \in \mathbb{R}^d$ with distribution $P_X$ that is absolutely continuous with respect to Lebesgue measure and whose support is not contained in any proper affine subspace of $\mathbb{R}^d$.

\begin{theorem}[Jet-based identifiability in linear two-module pipelines]
\label{thm:jet-identifiability-linear}
Consider two linear two-module pipelines without bias,
\[
f(x) = g \circ h(x) = w^\top H x,
\qquad
\tilde{f}(x) = \tilde{g} \circ \tilde{h}(x) = \tilde{w}^\top \tilde{H} x,
\]
with $H,\tilde{H} \in \mathbb{R}^{r \times d}$ and $w,\tilde{w} \in \mathbb{R}^r$, and a tap after the first module so that
\[
z(x) = h(x) = Hx, 
\qquad 
\tilde{z}(x) = \tilde{h}(x) = \tilde{H}x.
\]
Let $X$ be a random input with distribution $P_X$.

\begin{enumerate}[(a)]
    \item \emph{(Risk-only mirage family.)} If we only require that the composed maps coincide almost surely,
    \[
    w^\top H X = \tilde{w}^\top \tilde{H} X \quad \text{a.s.\ under } P_X,
    \]
    then for any invertible $Q \in \mathbb{R}^{r \times r}$ the reparameterised decomposition
    \[
    h_Q(x) := Q H x, 
    \qquad 
    g_Q(z) := w^\top Q^{-1} z
    \]
    defines a pipeline $f_Q = g_Q \circ h_Q$ with $f_Q(X) = f(X)$ almost surely. Thus, risk-only evaluation cannot distinguish among the $r^2$-parameter family $\{(g_Q,h_Q): Q \in \mathrm{GL}_r\}$ of mirage decompositions.
    
    \item \emph{(Jet-based identifiability.)} Suppose, in addition, that
    \begin{enumerate}[(i)]
        \item $f(X) = \tilde{f}(X)$ almost surely under $P_X$;
        \item the module-level jets at the tap coincide almost everywhere in the sense that
        \[
        J(x) = \tilde{J}(x) \quad \text{for $P_X$-a.e.\ } x,
        \]
        where $J(x)$ and $\tilde{J}(x)$ are the jets defined in~\eqref{eq:mojet} for the tapped module; 
        \item $H$ has full row rank $r$, and the support of $P_X$ is not contained in any proper affine subspace of $\mathbb{R}^d$.
    \end{enumerate}
    Then $H = \tilde{H}$ and $w = \tilde{w}$. In particular, among all linear two-module decompositions with the same input--output map, the one that matches the observed jets at the tap is unique (up to the trivial identity symmetry).
\end{enumerate}
\end{theorem}

\begin{proof}
For part (a), fix any invertible $Q \in \mathbb{R}^{r \times r}$ and define $h_Q$ and $g_Q$ as above. Then, for all $x \in \mathbb{R}^d$,
\[
g_Q \circ h_Q(x) = w^\top Q^{-1} (Q H x) = w^\top H x = f(x),
\]
so $f_Q(X) = f(X)$ almost surely under $P_X$. Distinct choices of $Q$ give distinct internal decompositions whenever $H$ has full row rank. 

For part (b), from $J(x) = \tilde{J}(x)$ for $P_X$-almost every $x$, we have $z(x) = \tilde{z}(x)$ and $A = \tilde{A}$ almost everywhere. Since $z(x) = Hx$ and $\tilde{z}(x) = \tilde{H}x$, and the support of $P_X$ is not contained in any proper affine subspace, the equality $Hx = \tilde{H}x$ for $P_X$-almost every $x$ implies $H = \tilde{H}$ as matrices. Using $H = \tilde{H}$ and the equality of composed maps,
\[
w^\top H x = \tilde{w}^\top H x \quad \text{for $P_X$-a.e.\ } x,
\]
we obtain
\[
(w - \tilde{w})^\top H x = 0 \quad \text{for $P_X$-a.e.\ } x.
\]
Because $H$ has full row rank $r$ and the support of $P_X$ is not contained in a proper affine subspace, the image $H X$ has support not contained in any proper affine subspace of $\mathbb{R}^r$. Thus the linear functional $(w - \tilde{w})^\top z$ vanishes almost surely on a set whose affine span is all of $\mathbb{R}^r$, forcing $w - \tilde{w} = 0$. Hence $w = \tilde{w}$ and the decomposition is uniquely pinned down by the jets.
\end{proof}
The theorem shows that, in this simple linear setting, module-level jets can lift identifiability from the level of the composed map to the level of the internal factorisation, whereas risk-only evaluation is insensitive to an entire $\mathrm{GL}_r$-orbit of mirage decompositions. The next corollary connects this idealised picture to the empirical jet estimator~\eqref{eq:jet-closed-form} used in practice.

\begin{corollary}[Consistency of linear jets under ideal perturbations]
\label{cor:linear-jets}
In the setting of Theorem~\ref{thm:jet-identifiability-linear}, fix a base point $x_0$ and suppose the tap is at the bottleneck, so that $z(x) = Hx$ with constant Jacobian $A = H$. Consider the empirical jet estimator~\eqref{eq:jet-closed-form} with $\lambda = 0$, and perturbations $\{\delta_j\}_{j=1}^J$ collected into $\Delta \in \mathbb{R}^{J \times d}$ as in Section~\ref{sec:jets}. Assume
\begin{enumerate}[(i)]
    \item the pipeline is noise-free and linear in $x$, so that $z(x_0 + \delta_j) - z(x_0) = H \delta_j$ for all $j$;
    \item $\Delta$ has full column rank $d$.
\end{enumerate}
Then the empirical jet recovers the true Jacobian exactly
\[
A(x_0) = H.
\]
In particular, under ideal perturbations and noiseless linear modules, module-level jets are exactly observable, and the identifiability statement of Theorem~\ref{thm:jet-identifiability-linear} applies directly.
\end{corollary}

\begin{proof}
By linearity and the noiseless assumption,
\[
Y_m \in \mathbb{R}^{J \times d_m} \quad \text{has rows} \quad
Y_{m,j,\cdot} = (z(x_0 + \delta_j) - z(x_0))^\top = (H \delta_j)^\top = \delta_j^\top H^\top,
\]
so $Y_m = \Delta H^\top$. With $\lambda = 0$, the empirical jet estimator~\eqref{eq:jet-closed-form} gives
\[
A(x_0)^\top = (\Delta^\top \Delta)^{-1} \Delta^\top Y_m
           = (\Delta^\top \Delta)^{-1} \Delta^\top \Delta H^\top
           = H^\top,
\]
where we used that $\Delta$ has full column rank and hence $\Delta^\top \Delta$ is invertible. Transposing both sides yields $A(x_0) = H$, as claimed.
\end{proof}

\section{MoJet: Empirical jets and mirage diagnostics}
\label{sec:algorithm}

\subsection{Algorithm}
\label{subsec:algorithm}

At a high level, Modular Jets turn a trained pipeline into a collection of \emph{local linear models} around selected base inputs. The algorithm does not look inside the modules beyond the ability to tap their representations; it only needs to (i) run the full pipeline on perturbed inputs and (ii) read out the tapped vectors. For each module $M_m$ and each base input $x^{(s)}$, we fit a ridge-regularised linear map that predicts changes in the tapped representation from small input perturbations. This map is the empirical Jacobian $A_m(x^{(s)})$ of the module at $x^{(s)}$, and the pair $(z_m(x^{(s)}), A_m(x^{(s)}))$ is the module's jet at that point. We refer to Algorithm~\ref{alg:jet-eval} as the \emph{Modular Jet Diagnostics} (MoJet) procedure. The input consists of a modular pipeline $x \mapsto z_1 \mapsto \cdots \mapsto z_K$; tap functions $\mathrm{tap}_m$ giving module representations; a set of base inputs $\{x^{(s)}\}$; a collection of perturbations $\{\delta_j\}$ that define the probe family (coarse or structured); and a ridge parameter $\lambda$. The output is a collection of empirical jets, numerical ranks, and jet similarity scores that can be used to diagnose mirage regimes. The algorithm has three conceptually distinct stages.

\begin{algorithm}[h]
\caption{Modular Jet Diagnostics (MoJet): Empirical Jet Estimation and Mirage Analysis}
\label{alg:jet-eval}
\begin{algorithmic}[1]
\Require Modular pipeline $x \mapsto z_1 \mapsto \cdots \mapsto z_K = \hat{y}$; tap functions $\mathrm{tap}_m$; base inputs $\{x^{(s)}\}_{s=1}^S$; perturbations $\{\delta_j\}_{j=1}^J$; ridge parameter $\lambda \ge 0$.
\Ensure Empirical jets $\{J_m(x^{(s)})\}$; jet similarity scores; numerical ranks.
\Statex
\State \textbf{Data collection.}
\For{$s = 1,\dots,S$}
    \State $x^{(s,0)} \gets x^{(s)}$
    \For{$j = 1,\dots,J$}
        \State $x^{(s,j)} \gets x^{(s)} + \delta_j$
        \State Run full pipeline on $x^{(s,j)}$
        \For{$m = 1,\dots,K$}
            \State $z_m^{(s,j)} \gets \mathrm{tap}_m(x^{(s,j)})$
        \EndFor
    \EndFor
\EndFor
\Statex
\State \textbf{Jet estimation.}
\For{$m = 1,\dots,K$}
    \For{$s = 1,\dots,S$}
        \State Form $\Delta \in \mathbb{R}^{J \times d}$ with rows $\delta_j^\top$
        \State Form $Y_m^{(s)} \in \mathbb{R}^{J \times d_m}$ with rows $(z_m^{(s,j)} - z_m^{(s,0)})^\top$
        \State Compute $A_m(x^{(s)})^\top \gets (\Delta^\top \Delta + \lambda I_d)^{-1} \Delta^\top Y_m^{(s)}$
        \State Set $J_m(x^{(s)}) \gets (z_m^{(s,0)}, A_m(x^{(s)}))$
    \EndFor
\EndFor
\Statex
\State \textbf{Diagnostics.}
\For{$m = 1,\dots,K$}
    \For{$s = 1,\dots,S$}
        \State Compute numerical rank $r_m^{(s)}$ from the singular values of $A_m(x^{(s)})$
    \EndFor
\EndFor
\ForAll{pairs $(m,m')$ with $m < m'$}
    \For{$s = 1,\dots,S$}
        \State Compute jet similarity $\mathrm{JetSim}(A_m(x^{(s)}), A_{m'}(x^{(s)}))$
    \EndFor
\EndFor
\State \Return $\{J_m(x^{(s)})\}$, $\{r_m^{(s)}\}$, and jet similarity scores.
\end{algorithmic}
\end{algorithm}
\paragraph{\emph{(A)} Data collection.}
For each base input $x^{(s)}$ and each perturbation $\delta_j$, we construct a perturbed input $x^{(s,j)} = x^{(s)} + \delta_j$ and run the \emph{entire} pipeline on it. At each tap $\mathrm{tap}_m$ we record the resulting representation $z_m^{(s,j)}$. This treats the trained pipeline as a black box with multiple observable outputs: we never differentiate the modules analytically or modify their internals.
\paragraph{\emph{(B)} Jet estimation via ridge regression.}
For a fixed module $m$ and base $x^{(s)}$, we view the pairs $\{(\delta_j, z_m^{(s,j)} - z_m^{(s,0)})\}_{j=1}^J$ as noisy samples from a local linear model in the input space. Stacking the perturbations into a matrix $\Delta$ and the response differences into $Y_m^{(s)}$, we solve a ridge-regularised least-squares problem to obtain the empirical Jacobian $A_m(x^{(s)})$ in closed form. The resulting matrix describes how the tapped representation changes, to first order, in response to small input perturbations drawn from the chosen probe family. The pair $J_m(x^{(s)}) = (z_m^{(s,0)}, A_m(x^{(s)}))$ is the empirical jet of module $m$ at $x^{(s)}$.

\paragraph{\emph{(C)} Diagnostics.}
Given the collection of Jacobians, we extract two simple summaries.
\begin{enumerate}
    \item Numerical rank $r_m^{(s)}$ of $A_m(x^{(s)})$, computed from its singular values with a relative tolerance. This measures the effective number of input directions to which module $m$ is locally sensitive at $x^{(s)}$ under the chosen perturbations.
    \item Jet similarity between two modules (or between the same module in two different models), implemented via subspace affinity between their input subspaces. In our experiments, $\mathrm{JetSim}(A,B)$ is computed by taking thin singular value decompositions of $A^\top$ and $B^\top$, extracting the leading singular vectors up to a relative threshold, and averaging the cosines of the principal angles between the resulting subspaces. The score lies in $[0,1]$, with $1$ indicating identical subspaces and $0$ indicating orthogonality.
\end{enumerate}
High jet similarity and low ranks across modules and base points indicate that, under the current perturbation design, the modules are locally indistinguishable as linear operators on input space; this is suggestive of a \emph{mirage regime}, where different decompositions behave the same way on all probed inputs and directions. By contrast, systematic differences in rank (for example, one bottleneck consistently using fewer effective degrees of freedom than another) or persistent low similarity between modules provide empirical evidence that the decomposition carries distinct functional roles which are constrained by the data and perturbations. In the synthetic experiments and the digits case study, these diagnostics are precisely what distinguish pipelines that look interchangeable under standard risk from those that differ in their internal geometry.

\subsection{Link to linear jet-identifiability}

In the linear, noiseless setting of Theorem~\ref{thm:jet-identifiability-linear}, Corollary~\ref{cor:linear-jets} shows that the empirical jet estimator~\eqref{eq:jet-closed-form} recovers the true module Jacobian exactly whenever the perturbation matrix $\Delta$ has full column rank. In other words, MoJet is an exact jet oracle in this idealised regime, and the identifiability guarantees of Theorem~\ref{thm:jet-identifiability-linear} apply directly. In our practical implementation, we use a small ridge term to regularise the local linear system; in the linear case, this converges to the same solution as the closed-form estimator as the number of probes grows and the ridge scale is kept small. 

In nonlinear pipelines, MoJet provides a controlled approximation to these ideal jets, where finite probes, local perturbations, and regularisation induce an approximation error; however, the resulting jets still capture the dominant local response directions. The diagnostics in Subsection~\ref{subsec:algorithm} and the robustness experiments in Section~\ref{sec:robustness-compute} then quantify how far two decompositions can be regarded as indistinguishable (a mirage) or empirically separated (identifiable) under a given perturbation design.

\section{Experiments}
\label{sec:experiments}

We now illustrate the Modular Jet perspective on four small but concrete supervised-learning setups. (i) Linear regression as a sanity check, (ii) a two-module deep regressor with different internal decompositions but similar risk, (iii) a synthetic pipeline classifier (PCA and logistic regression) contrasted with a monolithic dense classifier, and (iv) a real-data digits classification problem comparing a PCA and logistic pipeline to a one-hidden-layer MLP. The first three setups use low-dimensional synthetic data with known generative structure, while the digits experiment uses real handwritten digit images.\footnote{All models are implemented in PyTorch or scikit-learn and are intentionally kept modest in size, allowing jet behaviour to be visualised, interpreted, and stress-tested via small ablations.}

Throughout, we report the following. (i) Standard test risk (MSE or classification error), (ii) summary statistics of jet similarity between modules, averaged over base points, and (iii) distributions of numerical ranks of the empirical Jacobians $A_m(x^{(s)})$ for selected modules. For the synthetic experiments, we focus on how jets distinguish between models that have similar risk but different internal decompositions. For the digits experiment, we ask whether the same qualitative patterns appear when moving from synthetic to real data, and we additionally examine how MoJet’s similarity profiles behave under changes in perturbation scale, number of probes, and subspace dimensionality, as well as the associated compute cost.

Table~\ref{tab:summary} summarises the synthetic setups and their key metrics, and Figures~\ref{fig:linreg-jets}--\ref{fig:pca-logistic-jets} contain representative visualisations of jets and diagnostics for these three experiments. The real-data digits experiment has its own summary in Table~\ref{tab:digits-summary} and Figure~\ref{fig:digits-pca-mlp-jets}, with hyperparameter and subspace-sensitivity ablations in Figures~\ref{fig:hyperparam-sensitivity}--\ref{fig:k-sensitivity} and the corresponding compute overhead reported in Table~\ref{tab:compute-cost}.

\begin{table}[t]
\centering
\caption{Summary of synthetic experimental setups and metrics. Test risk is MSE for regression and classification error for classification. JetSim is the average subspace similarity between the relevant module jets (higher = more similar). Rank is the median numerical rank of the corresponding Jacobians.}
\label{tab:summary}
\scalebox{0.6}{
\begin{tabular}{llccc}
\toprule
\textbf{Experiment} & \textbf{Model(s)} & \textbf{Test risk} & \textbf{Avg.\ JetSim} & \textbf{Median rank} \\
\midrule
Linear regression &
$y = x^\top \beta + \varepsilon$ &
$0.0113$ &
-- &
$10$ \\
Two-module deep regressor &
$g \circ h$ vs.\ $\tilde{g} \circ \tilde{h}$ &
$0.0150 / 0.0221$ &
$0.68$ (coarse) / $0.84$ (structured) &
$3$ (A; both) / $1$ (B; coarse), $2$ (B; struct.) \\
Pipeline classification &
PCA and logistic vs.\ dense layer &
$0.1333 / 0.1367$ &
$0.44$ (coarse) / $1.00$ (aligned) &
$3$ (PCA; both) / $20$ (Dense; coarse), $3$ (Dense; aligned) \\
\bottomrule
\end{tabular}
}
\end{table}

\subsection{Linear regression sanity check}
\label{subsec:linreg}

\begin{figure}[t]
\centering
\includegraphics[width=0.7\textwidth]{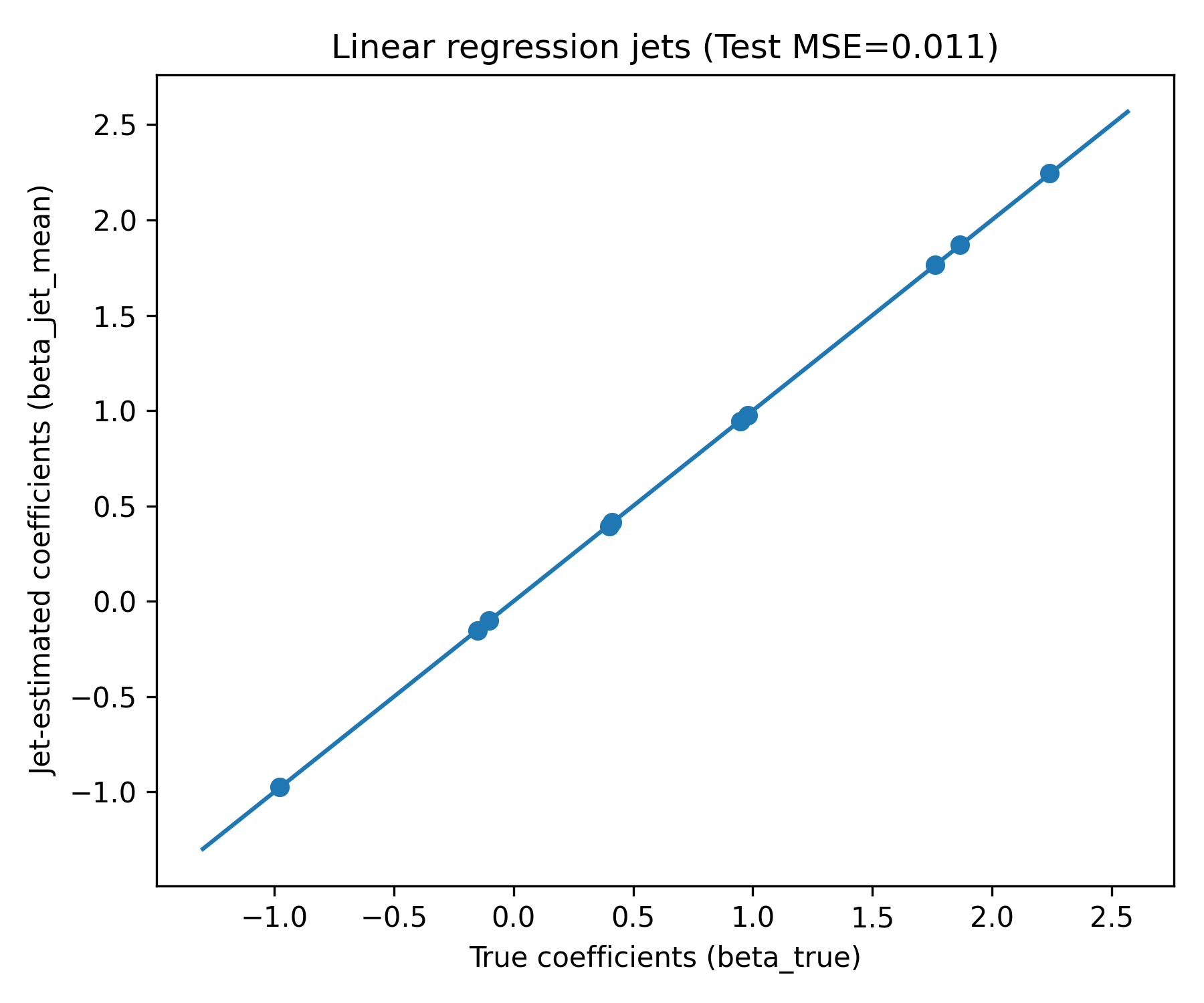}
\caption{Linear regression sanity check. Each point corresponds to one coordinate of the jet-estimated gradient $\hat{\beta}_{\mathrm{jet}}$ (averaged over base inputs) versus the corresponding ground-truth coefficient $\beta^\star$. The diagonal line indicates perfect agreement. The close clustering around the diagonal confirms that empirical jets recover the true linear parameters up to small numerical error.}
\label{fig:linreg-jets}
\end{figure}

We begin with a simple linear regression model
\[
Y = X^\top \beta^\star + \varepsilon,
\]
with $X \in \mathbb{R}^{d}$, $d=10$, and $\varepsilon \sim \mathcal{N}(0,\sigma^2)$. We sample $N_{\mathrm{train}} = 1000$ training points and $N_{\mathrm{test}} = 200$ test points from a Gaussian design $X \sim \mathcal{N}(0,I_d)$, and fit ordinary least squares without intercept to obtain $\hat{\beta}$. The learned predictor is $f(x) = x^\top \hat{\beta}$. On the run reported here, the test mean-squared error is $0.0113$. For $S=20$ base inputs $\{x^{(s)}\}$ drawn from the test set, we estimate the global jet
\[
J(x^{(s)}) = \bigl(f(x^{(s)}), A(x^{(s)})\bigr),
\]
where $A(x^{(s)})$ is the empirical Jacobian of $f$ with respect to $x$ at $x^{(s)}$, estimated via finite differences as in Section~\ref{sec:jets}. Because $f$ is globally linear, $A(x^{(s)})$ is constant in $x^{(s)}$ and coincides with the gradient $\nabla f = \hat{\beta}$. Empirically, the jet-estimated gradient $\hat{\beta}_{\mathrm{jet}}$ (averaged over base points) matches the ground-truth coefficients $\beta^\star$ up to small error. The coordinate-wise RMSE between $\beta^\star$ and $\hat{\beta}_{\mathrm{jet}}$ is approximately $3.4\times 10^{-3}$ which is comparable to the OLS estimator itself. Figure~\ref{fig:linreg-jets} plots, for each coordinate, the jet-estimated gradient against the true coefficient. In this globally linear model with no non-trivial modular decomposition, jets reduce to the familiar parameter vector and do not reveal additional structure. This serves as a sanity check for the empirical jet estimation routine.

\subsection{Two-module deep regressor}
\label{subsec:deep-regressor}

\begin{figure}[t]
\centering
\includegraphics[width=0.95\textwidth]{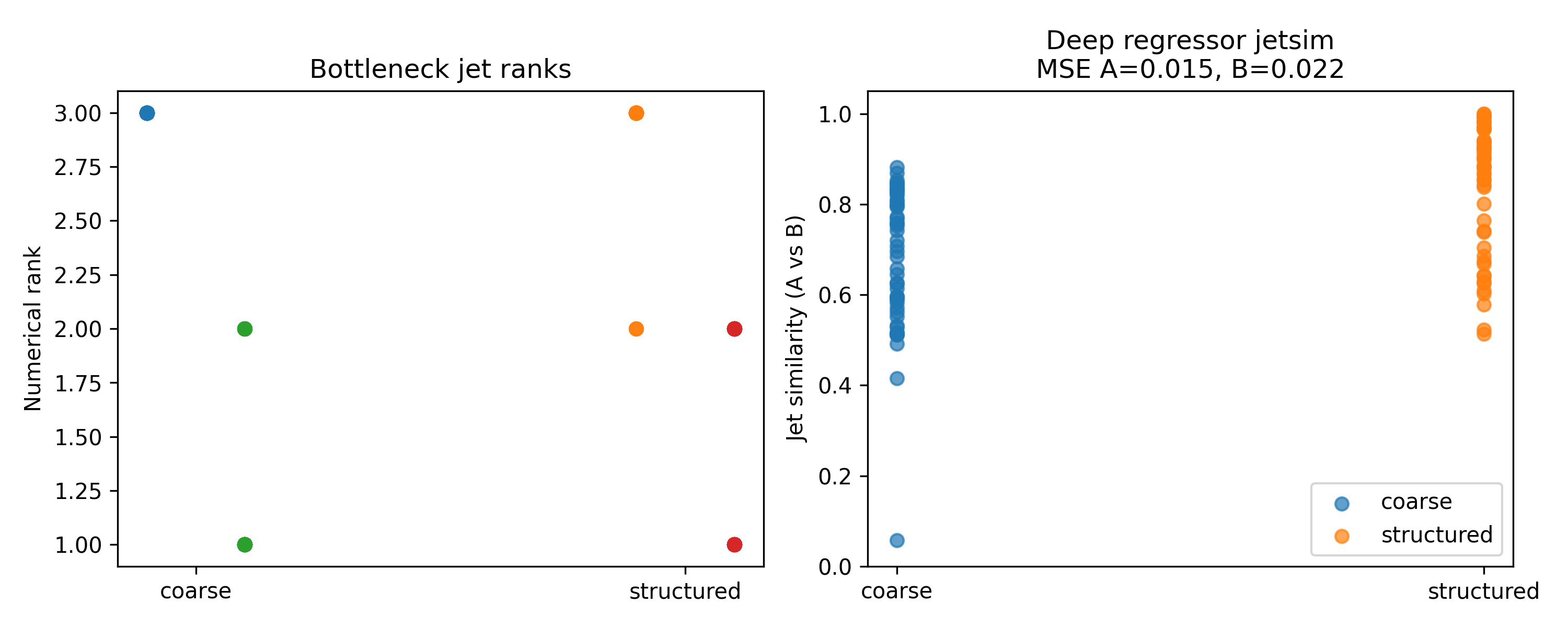}
\caption{Two-module deep regressor. Left: numerical ranks of bottleneck jets for Model~A and Model~B under coarse (left cluster) and structured (right cluster) perturbations. Model~A consistently has rank~$3$, while Model~B has rank close to~$1$ (coarse) and between~$1$ and~$2$ (structured). Right: jet similarity between the bottleneck modules of Model~A and Model~B under coarse and structured perturbations. Average similarity increases from about $0.68$ (coarse) to about $0.84$ (structured), indicating stronger alignment along task-relevant directions.}
\label{fig:deep-regressor-jets}
\end{figure}

We next consider a low-dimensional nonlinear regression problem designed to admit multiple internal decompositions with similar predictive performance. Inputs $x \in \mathbb{R}^8$ are generated from a Gaussian latent variable $z \in \mathbb{R}^{k}$ with $k=3$, embedded by a random linear map and corrupted by noise. The target $y$ is a smooth nonlinear function of $z$ (tanh plus a quadratic term), so the Bayes-optimal regressor depends only on a three-dimensional latent subspace. We train two networks on the same data.
\begin{enumerate}
    \item Model A (explicit bottleneck). A two-module network $f_A(x) = g_A(h_A(x))$ where $h_A: \mathbb{R}^8 \to \mathbb{R}^3$ is a bottlenecked feature extractor and $g_A: \mathbb{R}^3 \to \mathbb{R}$ is a small head.
    \item Model B (rotated bottleneck). A network $f_B(x) = \tilde{g}_B(\tilde{h}_B(x))$ with the same total capacity but additional linear layers before and after the bottleneck, creating a different internal decomposition while keeping overall expressivity comparable.
\end{enumerate}
Both networks are trained with MSE loss. On the test set, Model~A achieves MSE $0.0150$ and Model~B achieves $0.0221$, so from the perspective of risk they are close but not identical. We instrument taps at the bottleneck-like modules of both models (the output of $h_A$ and the effective bottleneck in Model~B). For $S=50$ base inputs from the test distribution and two perturbation families,
\begin{enumerate}[(a)]
    \item coarse isotropic noise $\delta \sim \mathcal{N}(0,\sigma^2 I)$, and
    \item structured perturbations formed by random combinations of the top $k=3$ principal components of $X$,
\end{enumerate}
we estimate bottleneck jets $J_1(x^{(s)})$ for each model, compute the numerical rank of the Jacobians, and the jet similarity between corresponding modules. The results can be summarised as follows (see Figure~\ref{fig:deep-regressor-jets} and Table~\ref{tab:summary}).
\begin{enumerate}
    \item \emph{Effective dimensionality}. For Model~A, the numerical rank of the bottleneck Jacobian is essentially three under both perturbation families (mean rank $3.00$ for coarse noise and $2.94$ for structured, with median $3$ in both cases). For Model B, the bottleneck Jacobian is of lower rank. Under coarse noise, the mean and median ranks are $1.18$ and $1$ respectively; under structured perturbations, the mean rank is $1.50$, with median between $1$ and $2$. Thus, even though both models achieve similar risk, Model~B compresses the input manifold more aggressively at the bottleneck.
    \item \emph{Jet similarity}. Despite this rank difference, the input subspaces to which the two bottlenecks are sensitive are moderately aligned. The average jet similarity $\mathrm{JetSim}(A_A, A_B)$ is approximately $0.68$ under coarse noise and increases to approximately $0.84$ under structured perturbations that follow the principal directions. The $75$th percentile of jet similarity under structured perturbations exceeds $0.94$, meaning that for many base points the two bottlenecks behave very similarly along task-relevant directions.
\end{enumerate}
From a Modular Jet perspective, these results illustrate a partial mirage. Standard risk-based evaluation suggests that the two decompositions are almost interchangeable, and even their jets are substantially aligned, especially along principal directions. At the same time, the systematically lower rank of Model~B's bottleneck jets indicates that its internal representation is genuinely different. It uses fewer effective degrees of freedom to span the same latent subspace. Jet-based diagnostics, therefore, add information that is not visible from risk alone.

\subsection{Pipeline classification: PCA and logistic regression}
\label{subsec:pipeline-classification}

\begin{figure}[t]
\centering
\includegraphics[width=0.95\textwidth]{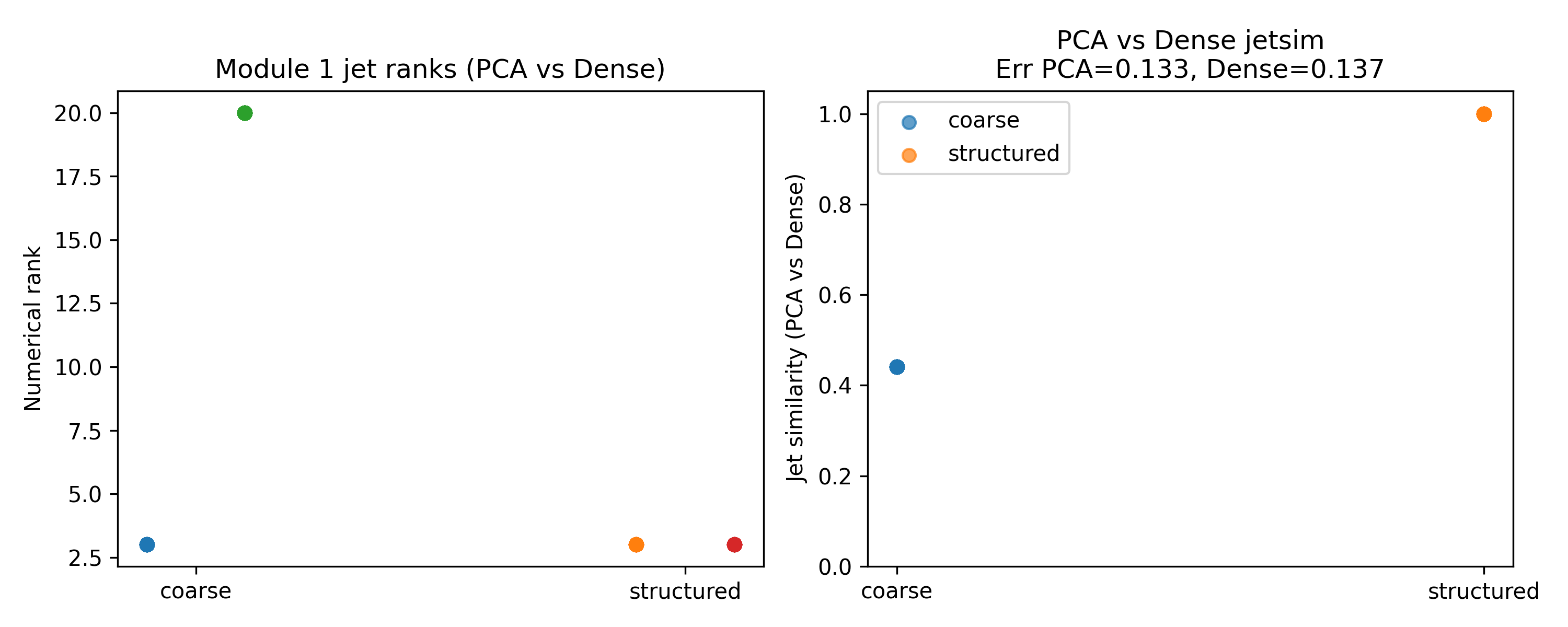}
\caption{Pipeline classification (PCA and logistic regression vs.\ dense classifier). Left: numerical ranks of module~1 jets under coarse and structured perturbations. The PCA module is always rank~$3$; the dense model is rank~$20$ under coarse noise but collapses to rank~$3$ when perturbations are restricted to the principal subspace. Right: jet similarity between the PCA module and the dense model's identity module. Similarity is low under coarse perturbations ($\approx 0.44$) but essentially~$1.00$ under aligned perturbations, indicating that the two pipelines behave identically along task-relevant directions but differ sharply in their treatment of nuisance directions.}
\label{fig:pca-logistic-jets}
\end{figure}

Our next example considers a classification pipeline with an explicit dimensionality-reduction stage. Inputs $x \in \mathbb{R}^{20}$ and labels $y \in \{1,2,3\}$ are generated by a latent mixture model: we sample a class-dependent latent vector $z^\star \in \mathbb{R}^3$, embed it into $\mathbb{R}^{20}$ via a random linear map, and add Gaussian noise. The Bayes-optimal classifier depends only on the three-dimensional latent subspace. We train two classifiers.
\begin{enumerate}
    \item Pipeline model (PCA and logistic). A modular pipeline with $M_1$ performing PCA to dimension $k=3$, followed by $M_2$ performing multinomial logistic regression on the principal components.
    \item Dense logistic model. A single multinomial logistic regression layer trained directly on the original $20$-dimensional inputs, with no explicit dimensionality reduction.
\end{enumerate}
On the test set, the PCA and logistic pipeline and the dense classifier achieve classification errors of $0.1333$ and $0.1367$, respectively, which are essentially indistinguishable at the level of standard evaluation. We instrument taps at the intermediate representation in both models: the PCA scores for the pipeline and the raw input (identity map) for the dense model. For $S=50$ base inputs from the test distribution, we consider the following. 
\begin{enumerate}[(a)]
    \item Coarse perturbations. Small isotropic Gaussian noise around each base input.
    \item Aligned perturbations. Perturbations are restricted to the top three principal directions, so that all perturbations lie in the task-relevant subspace.
\end{enumerate}
For each base input, model, and perturbation family, we estimate the jet of module~1 and compute its numerical rank and jet similarity across models. The resulting picture is (see Figure~\ref{fig:pca-logistic-jets} and Table~\ref{tab:summary}).
\begin{enumerate}[(a)]
    \item \emph{Coarse perturbations}. The jet of the PCA module has numerical rank $3$ for all bases, as expected. The jet of the dense model's identity module has rank $20$ under coarse perturbations (mean and median both $20$). The average jet similarity between the two modules under coarse noise is about $0.44$, indicating that, as linear maps on the input, they respond quite differently to generic perturbations: the dense model is sensitive along many nuisance directions in addition to the task-relevant subspace.
    \item \emph{Aligned perturbations}. When perturbations are restricted to the principal subspace, both modules have numerical rank $3$, and the average jet similarity jumps to essentially $1.00$ (mean $0.99999$). Along task-relevant directions, the dense classifier's representation behaves almost identically to explicit PCA followed by logistic regression; the difference between the pipelines lies almost entirely in how they treat directions orthogonal to that subspace.
\end{enumerate}

Taken together, these findings show how Modular Jets make explicit the dependence on evaluation design. If we only probe along the principal subspace, the two pipelines appear functionally identical at the representation level (jet similarity $\approx 1$). If we probe in the full ambient space, the dense model reveals a much higher effective dimensionality and a very different sensitivity profile (jet similarity $\approx 0.44$). Both models have nearly the same classification error, but their internal geometry with respect to nuisance directions is sharply different.

\subsection{Digits classification with PCA and MLP}
\label{subsec:digits-real}

\begin{figure}[t]
\centering
\includegraphics[width=0.95\textwidth]{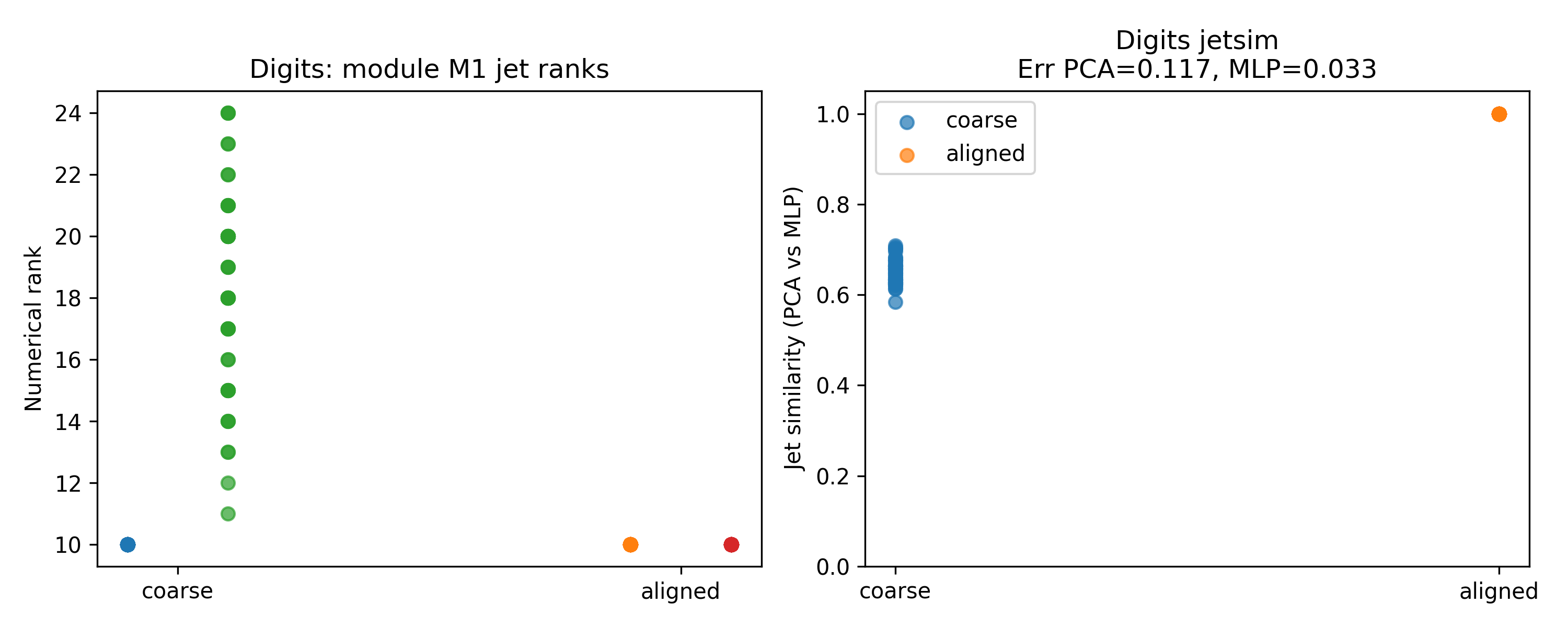}
\caption{Digits classification (PCA and logistic regression vs.\ one-hidden-layer MLP). Left: numerical ranks of module-1 jets under coarse and aligned perturbations for both models. The PCA module has rank~$10$ by construction; the MLP hidden layer exhibits higher rank under coarse perturbations (median $18$) but collapses to rank~$10$ under aligned perturbations. Right: jet similarity between the PCA module and the MLP hidden layer. Similarity is moderate under coarse perturbations (mean $0.649996$) but essentially~$1.0$ under aligned perturbations (mean $0.999986$), indicating that the two pipelines behave almost identically along task-relevant directions while differing in their response to nuisance directions.}
\label{fig:digits-pca-mlp-jets}
\end{figure}

To complement the synthetic setups, we conduct a small real-data experiment on the scikit-learn digits dataset, which contains $1797$ grayscale images of handwritten digits ($0$--$9$), each of size $8\times 8$ (flattened to $d=64$ features). Labels take values in $\{0,\dots,9\}$, and we use an 80/20 train--test split stratified by class. In a representative run, both models considered below achieve test misclassification error of roughly $0.11$, so standard evaluation would regard them as essentially equivalent. We compare two pipelines.
\begin{enumerate}
    \item \emph{PCA and logistic regression.} Inputs are standardised, then passed through PCA with $k=10$ components, followed by a multinomial logistic regression classifier on the principal scores.
    \item \emph{One-hidden-layer MLP.} Inputs are standardised, then fed into a one-hidden-layer neural network with hidden dimension $32$ and ReLU activation, followed by a linear layer to $10$ logits. The network is trained with cross-entropy loss in PyTorch.
\end{enumerate}
Both pipelines operate in the same scaled input space, but differ in how they implement dimensionality reduction and classification. We instrument taps at the first non-trivial module of each pipeline.
\begin{enumerate}[(a)]
    \item For the PCA and logistic pipeline, $\mathrm{tap}_1(x)$ returns the $k$-dimensional vector of PCA scores.
    \item For the MLP, $\mathrm{tap}_1(x)$ returns the hidden-layer representation (post-activation).
\end{enumerate}
Using the procedure of Section~\ref{sec:jets}, we estimate empirical jets $J_1(x^{(s)})$ at these taps for $S$ base inputs chosen from the test set, under two perturbation families in the scaled input space.
\begin{enumerate}[(a)]
    \item Coarse perturbations. Isotropic Gaussian noise $\delta \sim \mathcal{N}(0,\sigma^2 I_{64})$.
    \item Aligned perturbations. Perturbations restricted to the principal subspace, obtained as random linear combinations of the top $k$ PCA components.
\end{enumerate}
For each base input, model, and perturbation family, we compute (i) the numerical rank of the module-1 Jacobian, and (ii) the jet similarity between the PCA and MLP modules. Table~\ref{tab:digits-summary} summarises the resulting ranks and jet similarity scores while Figure~\ref{fig:digits-pca-mlp-jets} visualises their distributions. Two patterns emerge. 

First, the PCA module behaves as designed. Its module-1 Jacobian has numerical rank $10$ under both perturbation families. By contrast, the MLP hidden layer has substantially higher rank under coarse perturbations (median rank $18$, mean rank $18.16$), reflecting sensitivity to many directions in the ambient input space, but collapses to rank $10$ when perturbations are restricted to the principal subspace. Second, the jet similarity between the PCA module and the MLP hidden layer is only moderate under coarse perturbations (average similarity $0.649996$, median $0.647625$), but rises to essentially $1.0$ under aligned perturbations (average similarity $0.999986$). Along task-relevant directions, the MLP's hidden representation therefore behaves almost identically to the PCA scores; the difference between the pipelines lies primarily in how they respond to nuisance directions.

The digits experiment reproduces, on real data, the qualitative behaviour observed in the synthetic PCA pipeline: standard test error suggests that the PCA and logistic classifier and the MLP are interchangeable, but Modular Jets reveal that this equivalence is conditional on the perturbation family. If we only probe along the principal subspace, the two representations appear identical; if we probe in the full input space, they exhibit different effective dimensionalities and distinct local geometries.

\begin{table}[t]
\centering
\caption{Digits classification (real data). Jet diagnostics for the PCA and logistic pipeline and the one-hidden-layer MLP. ``Median rank'' is the median numerical rank of the module-1 Jacobians over base inputs. ``Avg.\ JetSim'' is the average jet similarity between the PCA scores module and the MLP hidden layer for the given perturbation family.}
\label{tab:digits-summary}
\scalebox{0.9}{
\begin{tabular}{lccc}
\toprule
\textbf{Perturbation} & \textbf{Model} & \textbf{Median rank (M1)} & \textbf{Avg.\ JetSim (PCA vs MLP)} \\
\midrule
Coarse (isotropic) &
PCA and logistic &
$10$ &
- \\
Coarse (isotropic) &
MLP (hidden layer) &
$18$ &
$0.649996$ \\
Aligned (top-$k$ PCs) &
PCA and logistic &
$10$ &
- \\
Aligned (top-$k$ PCs) &
MLP (hidden layer) &
$10$ &
$0.999986$ \\
\bottomrule
\end{tabular}
}
\end{table}

\subsection{Robustness to Hyperparameters and Compute Cost}
\label{sec:robustness-compute}

In this section, we provide a minimal set of ablations to demonstrate that our conclusions are not an artefact of fragile hyperparameter choices. Unless otherwise stated, all results are reported for the PCA module and the hidden layer of the MLP in the digits experiment.

\begin{figure}[t]
  \centering
  \begin{minipage}[t]{0.48\textwidth}
    \centering
    \includegraphics[width=\linewidth]{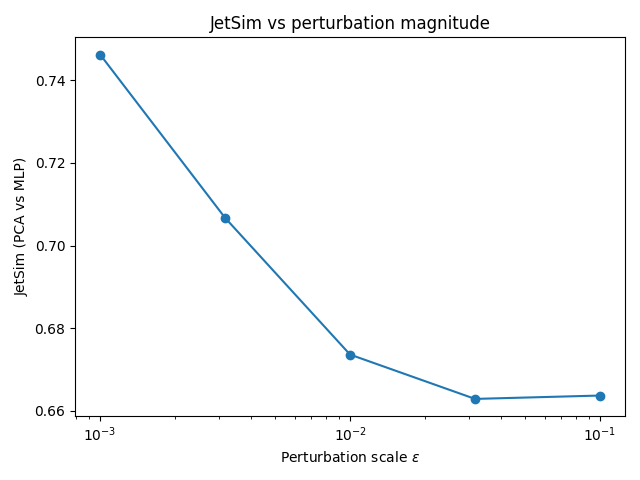}
    \vspace{-0.5em}
    \caption*{(a) JetSim vs perturbation magnitude $\varepsilon$ (log-scale).}
  \end{minipage}\hfill
  \begin{minipage}[t]{0.48\textwidth}
    \centering
    \includegraphics[width=\linewidth]{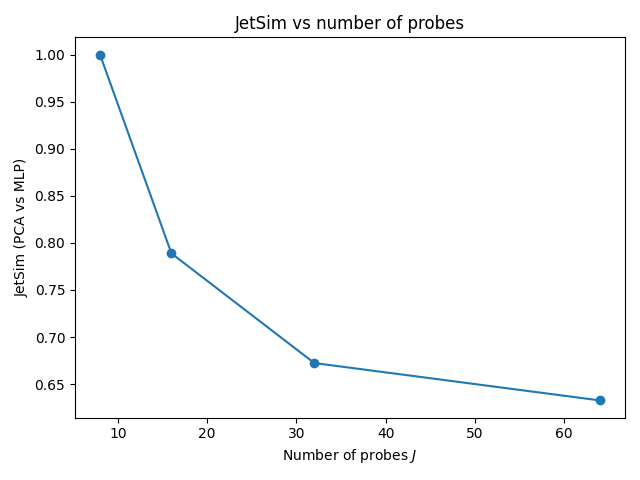}
    \vspace{-0.5em}
    \caption*{(b) JetSim vs number of probes $J$.}
  \end{minipage}
  \caption{Hyperparameter sensitivity for the PCA and MLP hidden modules on the digits task. Left: JetSim changes only mildly across more than two orders of magnitude in $\varepsilon$ and stabilises for $\varepsilon \gtrsim 10^{-2}$. Right: increasing $J$ beyond $32$ yields only small changes in JetSim, indicating diminishing returns once jets are well-conditioned.}
  \label{fig:hyperparam-sensitivity}
\end{figure}

\paragraph{Sensitivity to Perturbation Scale, Number of Probes, and Ridge.} We first study the effect of the perturbation magnitude $\varepsilon$ and the number of probes $J$ used in MoJet. For a fixed module and a fixed number of probes $J=32$, we sweep $\varepsilon$ over a log-spaced grid
\[
  \varepsilon \in \{10^{-3},\,10^{-3.5},\,10^{-2},\,10^{-1.5},\,10^{-1}\},
\]
and compute the resulting JetSim score between the PCA and MLP jets.
Figure~\ref{fig:hyperparam-sensitivity} (left) shows that JetSim varies only mildly across this range: it decreases from approximately $0.75$ at $\varepsilon = 10^{-3}$ to about $0.66$ for $\varepsilon \ge 10^{-2}$, and then stabilises. This suggests a broad regime of perturbation scales where jets are well-estimated but not dominated by floating-point noise. Next, we fix a representative perturbation scale $\varepsilon = 10^{-2}$ inside this stable region and vary the number of probes
\[
  J \in \{8, 16, 32, 64\}.
\]
As shown in Figure~\ref{fig:hyperparam-sensitivity} (right), increasing $J$ from $8$ to $16$ and $32$ substantially changes the estimated similarity (JetSim drops from $\approx 1.00$ at $J=8$ to $\approx 0.79$ and $\approx 0.67$, respectively), reflecting better-conditioned jets that no longer overfit a small set of probes. Beyond $J=32$ the curve flattens (JetSim $\approx 0.63$ at $J=64$), indicating diminishing returns. We therefore use $J=32$ as a practical trade-off between stability and cost in the main experiments. For the ridge parameter $\lambda$, we regularise the local linear system
\[
  \hat A = \arg\min_A \bigl\|Y - A U\bigr\|_F^2 + \lambda \|A\|_F^2,
\]
where columns of $U$ are probes (here analogous to the rows of $\Delta$ in Section~\ref{sec:jets}) and columns of $Y$ are corresponding responses. In all experiments we set $\lambda = \alpha \,\sigma_{\max}^2$, where $\sigma_{\max}^2$ is the largest eigenvalue of $U U^\top / J$ and we use $\alpha = 10^{-3}$. This scale-aware choice keeps the regularisation small relative to the dominant directions of $U U^\top$, and we did not observe any qualitative change in JetSim or jet ranks when varying $\alpha$ within the same order of magnitude.

\begin{figure}[t]
  \centering
  \includegraphics[width=0.6\linewidth]{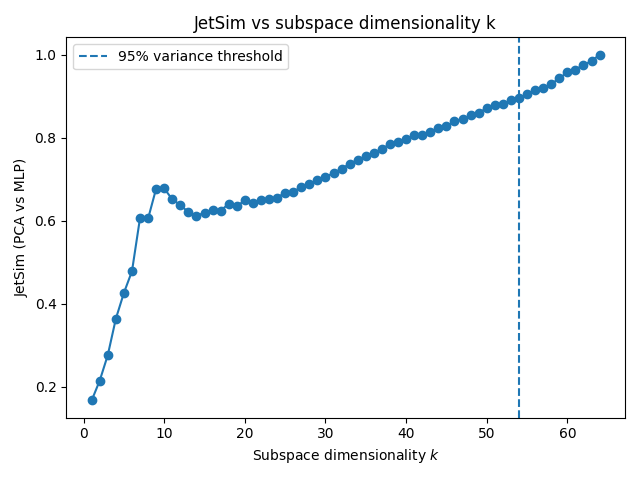}
  \caption{JetSim as a function of subspace dimensionality $k$ for the PCA and MLP jets on the digits task. The vertical dashed line indicates the $k=54$ chosen by the $95\%$ variance rule. JetSim continues to increase smoothly beyond the spectral ``elbow'', and remains high for $k$ in the range $[40,60]$, showing that our similarity conclusions are not sensitive to the exact cutoff.}
  \label{fig:k-sensitivity}
\end{figure}

\paragraph{Choice of Subspace Dimensionality for JetSim.} JetSim compares dominant jet subspaces via a principal-angle / projection-overlap metric. To choose the subspace dimensionality $k$, we consider the eigenspectrum of the empirical input-space covariance of the PCA jets,
\[
  C = \frac{1}{N} A^\top A,
\]
and retain the smallest $k$ such that a fixed fraction of the variance is captured $\frac{\sum_{i=1}^k \lambda_i}{\sum_{i=1}^d \lambda_i} \geq \rho$,
with $\rho = 0.95$ in all experiments. Figure~\ref{fig:k-sensitivity} shows JetSim as a function of $k$. The curve rises from $\approx 0.17$ at $k=1$ to around $0.66$ by $k \approx 10$, dips slightly, and then increases steadily, crossing $0.8$ near $k \approx 40$ and reaching $\approx 0.90$ at $k=54$. The vertical dashed line marks $k=54$, which is the smallest value achieving the $95\%$ variance threshold. Using smaller or larger $k$ in the range $k \in [40,60]$ does not change the qualitative conclusion that the PCA and MLP jets share a highly aligned dominant subspace.

\paragraph{Compute Cost.} We report the computational cost of MoJet on the digits task in Table~\ref{tab:compute-cost}. For each configuration, we measure the number of forward passes and wall-clock time (on a single CPU), counting only the additional overhead of MoJet relative to ordinary inference. As expected from the design, the cost scales linearly with the number of tapped modules $M$, the number of probed inputs $N_{\text{jet}}$, and the number of probes $J$ as
\[
  \text{\#fwd passes} \approx N_{\text{jet}} \times J \times M.
\]
In this experiment, we tap two modules (PCA and the MLP hidden layer), and probe $N_{\text{jet}} = 200$ inputs. Increasing $J$ from $32$ to $64$ doubles the number of forward passes and roughly doubles the wall-clock time, while leaving the qualitative JetSim profile unchanged.

\begin{table}[t]
  \centering
  \caption{Compute cost of MoJet on the digits experiment. Times are measured on a single CPU and reflect only the MoJet overhead, not training.}
  \label{tab:compute-cost}
  \vspace{0.5em}
  \begin{tabular}{lccccc}
    \toprule
    Configuration & Inputs $N_{\text{jet}}$ & Probes $J$ & Taps $M$ & Time (s) & \#fwd passes \\
    \midrule
    Digits (PCA \& MLP) & 200 & 32 & 2 & 6.29 & 12{,}800 \\
    Digits (PCA \& MLP) & 200 & 64 & 2 & 10.88 & 25{,}600 \\
    \bottomrule
  \end{tabular}
\end{table}


\paragraph{Non-smooth Modules and ReLU Networks.} Most of our modules are ReLU-based and hence piecewise affine. On each activation region, the mapping is exactly linear, so the module Jacobian is well-defined and constant almost everywhere; discontinuities only occur on measure-zero boundaries where neurons switch on or off. Finite-difference MoJet estimates may occasionally cross such boundaries, in which case the empirical Jacobian reflects a mixture of neighbouring regions. We mitigate this in two ways. (i) We use moderate perturbation scales $\varepsilon$ that remain in the local-linear regime but are not dominated by floating-point noise, and (ii) we average over multiple random probes. Empirically, JetSim varies smoothly across nearby inputs and hyperparameters (Figures~\ref{fig:hyperparam-sensitivity}--\ref{fig:k-sensitivity}), suggesting that boundary effects are effectively averaged out and do not drive our conclusions.

\subsection{Summary}
Across these examples, standard risk-based evaluation regards the compared models as essentially equivalent: linear regression is trivial, the two deep regressors have similar MSE, and the two classifiers have nearly identical error rates. Modular Jet analysis, even in this simple empirical form, reveals differences in internal structure: in the deep regressor, one bottleneck consumes fewer effective degrees of freedom; in the classification pipeline, the dense model and the PCA pipeline are indistinguishable along a high-variance principal subspace but diverge strongly in how they handle lower-variance (nuisance) directions. Our ablations show that these conclusions are stable across a broad range of perturbation scales, probe counts, and subspace dimensionalities, and can be obtained with a modest post-hoc compute overhead. In short, jets expose when similar risks hide different geometries, and when evaluation perturbations are too narrow to distinguish a genuine internal difference from a mirage.

\section{Discussion and Conclusion}
\label{sec:conclusion}

Risk-based evaluation is central to classical machine learning, but it provides a limited view of learned systems. It tells us which functions achieve low loss on a distribution, but not whether the internal decomposition of a model into modules is uniquely constrained by the data and evaluation design, or whether multiple distinct internal mechanisms remain observationally indistinguishable.

We have introduced Modular Jets as a simple tool to address this gap in the context of regression and classification. By treating the input space as a task manifold, instrumenting intermediate modules, and estimating local linear response maps under structured perturbations, we obtain empirical jets that characterise how each module behaves around sampled inputs. Comparing jets across modules and alternative decompositions provides a notion of mirage versus identifiability, which complements standard test risk. In an idealised linear two-module setting, we proved that access to module-level jets is sufficient to pin down the internal factorisation, while risk-only evaluation admits a large family of mirage decompositions that implement the same input--output map. The MoJet procedure can be viewed as a practical estimator of these jets: in the linear case, it recovers them exactly under mild rank conditions, and in nonlinear pipelines it provides a controlled approximation whose similarity profiles are empirically stable across a broad range of perturbation scales, probe counts, and subspace dimensionalities, with modest post-hoc compute overhead.

The Modular Jet perspective helps articulate an important question. \emph{To what extent do our evaluation designs constrain not just performance, but the internal structure of our models?} We see several directions for further work. Sharpening the empirical diagnostics into formal conditions beyond the linear case, extending the framework to multi-task or multi-modal models, and integrating jet-based analysis with existing notions of invariance, disentanglement, and interpretability~\cite{doshi2017towards,rudin2019stop,scholkopf2021toward,vonkugelgen2021nonparametric}. A complementary direction is to enrich the perturbation families (task-aware or semantic directions) and to turn descriptive mirage scores into formal hypothesis tests over jet subspaces. Ultimately, similar ideas can be extended beyond classical supervised learning to modern large-scale systems, including language models and more general AI agents, where questions of identifiability and mirage behaviour are even more pressing, and where component-level auditing and mechanistic analysis frameworks are beginning to emerge~\cite{semanticlens2025}.

We close by noting some limitations of the present analysis. Conceptually, the formal identifiability result is confined to linear two-module pipelines; in nonlinear and high-capacity networks, Modular Jets provide empirical diagnostics rather than guarantees, and we do not claim global identifiability beyond the linear setting. Methodologically, jets capture only local first-order behaviour under a chosen perturbation family; higher-order effects, non-smooth phenomena (ReLU kink points), and distribution shift beyond the task manifold are not modelled. Empirically, our experiments are deliberately small-scale, focusing on low- to medium-dimensional regression and classification rather than state-of-the-art architectures.  Nevertheless, these constraints make the examples interpretable, allowing us to isolate the core point: even very simple pipelines already exhibit mirage regimes that are invisible to risk-only evaluation, and module-level jets are a viable and practically robust method for probing them.

\end{document}